\documentclass[runningheads]{llncs}
\usepackage{graphicx}
\usepackage[utf8]{inputenc}

\usepackage{color, soul}

\usepackage{acronym}
\newacro{ot}[OT]{Operational Technology}
\newacro{ai}[AI]{Artificial Intelligence}
\newacro{xai}[XAI]{Explainable Artificial Intelligence}
\newacro{ers}[ERS]{Explainable Recommendation System}
\newacro{dnn}[DNN]{Deep Neural Network}

\usepackage{booktabs}

\usepackage{longtable}

\begin{document}

\title{On Explainability in AI-Solutions:\\A Cross-Domain Survey\thanks{This is a pre-print of an invited paper published in the Computer Safety, Reliability, and Security. SAFECOMP 2022 Workshops. Please cite as: \\ \textit{S.D. Duque Anton, D. Schneider, H.D. Schotten. On Explainability in AI-Solutions: A Cross-Domain Survey. In: Computer Safety, Reliability, and Security. SAFECOMP 2022 Workshops, SAFECOMP 2022, vol 13415. Springer, September 2022, DOI: 10.1007/978-3-031-14862-0\textunderscore 17}}}
\titlerunning{On Explainability in AI-Solutions}

\author{Simon D Duque Anton\inst{1}\orcidID{0000-0003-4005-9165} \and
Daniel Schneider\inst{2}\orcidID{0000-0002-1736-2417} \and
Hans D Schotten\inst{2}\orcidID{0000-0001-5005-3635}
}
\authorrunning{S. D. Duque Anton}
%
\institute{comlet Verteilte Systeme GmbH, 66482 Zweibruecken, Germany
\email{simon.duque-anton@comlet.de} \and
DFKI, 67663 Kaiserslautern, Germany \\
\email{\{Daniel, Hans\_Dieter\}.\{Schneider, Schotten\}@dfki.de}\\
}
\maketitle              
\begin{abstract}
Artificial Intelligence (AI) increasingly shows its potential to outperform predicate logic algorithms and human control alike.
In automatically deriving a system model,
AI algorithms learn relations in data that are not detectable for humans.
This great strength,
however,
also makes use of AI methods dubious.
The more complex a model,
the more difficult it is for a human to understand the reasoning for the decisions.
As currently,
fully automated AI algorithms are sparse,
every algorithm has to provide a reasoning for human operators.
For data engineers,
metrics such as accuracy and sensitivity are sufficient.
However,
if models are interacting with non-experts,
explanations have to be understandable. \par
This work provides an extensive survey of literature on this topic,
which,
to a large part,
consists of other surveys.
The findings are mapped to ways of explaining decisions and reasons for explaining decisions.
It shows that the heterogeneity of reasons and methods of and for explainability lead to individual explanatory frameworks.
\keywords{Artificial Intelligence \and Explainability \and Survey \and Cross-Domain.}
\end{abstract}

\section{Introduction}
Industrial revolutions provided humanity with novel technologies that fundamentally changed fields of work,
mostly in manufacturing and processing industries.
Currently,
the fourth industrial revolution is said to introduce flexibility and ad-hoc connectivity to once inflexible industrial \ac{ot} networks.
This newly integrated means of connectivity in industrial networks,
across physical factory boundaries,
allows for new use cases and improved efficiency.
Apart from the connectivity aspect,
the introduction of \ac{ai} methods presents new paradigms.
In industrial environments,
\ac{ai} methods are applied to production and resource planning~\cite{Elbasheer.2022},
detection of anomalies in production processes~\cite{Duque_Anton.2021,Duque_Anton.2019,Duque_Anton.2018},
and improvement of processes~\cite{Sun.2019}.
Apart from industrial applications,
\ac{ai} methods lend themselves readily on other domains,
such as finance and banking~\cite{Cao.2022},
medicine~\cite{Hamet.2017} and elderly care~\cite{Qian.2021},
autonomous driving~\cite{Schutera.2020,Schutera.2018}
network management~\cite{Shen.2020,Jiang.2019,Jiang.2017,Duque_Anton.1993},
and for control of unmanned vehicles~\cite{Sands.2020,Keneni.2019},
just to name a few.
In all of these fields,
automation and \ac{ai} are intended to perform tedious and repetitive tasks to relieve workers.
However,
this autonomous performance of tasks requires trustworthy and understandable algorithms as an enabler.
If a task is to be performed by an algorithm,
the outcome must not deviate from expectations,
jitters in the input data cannot change the outcome in an undesirable fashion.
Especially regarding \ac{ai} algorithms,
understanding the reasoning behind a decision is complex and often hardly possible for human operators.
This is an issue,
especially regarding regulatory standards and acceptance issues of users.
Consequently,
\ac{ai} algorithms need to be understandable and provide predictable outcomes in a reliable fashion to further their application.
This need has created the term of \ac{xai} that encompasses the need for \ac{ai} methods to not only provide sound results,
but also provide the reasoning in a useful and understandable manner.\\ \par
This work aims at providing an overview of requirements as well as solutions for the explainability of \ac{ai} algorithms. 
Works related to explainability in \ac{ai} are discussed in Section~\ref{sec:related_work}.
Methods and techniques for explaining outcomes of \ac{ai} algorithms are introduced in Section~\ref{sec:explainability_in_ai}.
Common application scenarios that require explainable \ac{ai} methods are presented in Section~\ref{sec:ai_scenarios}.
This work is concluded in Section~\ref{sec:conclusion}.

\section{Related Work}
\label{sec:related_work}
This section captures an overview of related works discussing explainability of \ac{ai} methods in different domains.
A comprehensive overview is provided in Table~\ref{tab:related_work_table}.
\begin{table}[ht!]
\renewcommand{\arraystretch}{1.3}
\caption{Overview of Related Works}
\label{tab:related_work_table}
\centering
\scriptsize
\begin{tabular}{r l l l}
\toprule
\textbf{Year} & \textbf{Work} & \textbf{Domain} & \textbf{Topic of Explainability} \\
 \cmidrule{1-4}
 2022 & \cite{Reddy.2022} & Medicine & Accuracy vs. explainability \\
 2021 & \cite{Neugebauer.2021} & Product \& company success  & Before and after model comparison \\
 2021 & \cite{Vilone.2021} & Survey & Comparison of methods aimed at human-based \\
  & & & understanding and objective metrics \\
 2021 & \cite{Caro.2021} & e-commerce & Ontological model \\
 2021 & \cite{Liang.2021} & Recommendation systems & Leverage of recommendation and explanation \\
 2021 & \cite{Holzinger.2021} & Medical domain & ``Mapping of explainability with causability'' \\
 2021 & \cite{Ehsan.2021} & Socio-economic decisions & Social transparency as guidelines for decision making  \\
 2021 & \cite{Angelov.2021} & Survey & Global vs. local, accuracy vs. explainability \\
 2021 & \cite{Mohseni.2021} & Survey & Multi-domain model \\
 2021 & \cite{Belle.2021} & Data science & Survey based on stakeholder use-case \\
 2021 & \cite{Shin.2021} & Product recommendation & Relationship of causability and explainability \\
 2020 & \cite{Singh.2020} & Medicine & Explainable image processing \\
 2020 & \cite{Barredo_Arrieta.2020} & Survey & Fairness of the model \\
 2020 & \cite{Lundberg.2020} & Tree-based algorithms & Game theory-based approach \\
 2020 & \cite{Ploug.2020} & Medicine & Contestability \\
 2020 & \cite{Arya.2020,Arya.2019} & Data Science & Toolkit-framework combining state of the art methods \\
 2020 & \cite{Linardatos.2020} & Survey \& Taxonomy & Survey \\
 2020 & \cite{Tjoa.2020} & Medicine & Survey \\
 2020 & \cite{Roscher.2020} & Natural sciences & Survey \\
 2020 & \cite{Beaudouin.2020} & Multidisciplinary & Three-step framework \\
 2020 & \cite{Coeckelbergh.2020} & General & Responsibility of \ac{ai} algorithms \\
 2020 & \cite{Bhatt.2020} & Survey & Stakeholder-driven explainability survey \\
 2020 & \cite{Ammar.2020} & Mental health evaluation & Ontological agent \\
 2020 & \cite{Zhang.2020} & Survey & Survey \\
 2020 & \cite{Confalonieri.2020} & Survey & Survey \\
 2020 & \cite{Amann.2020} & Medicine & Multidisciplinary survey of requirements in explainability \\
 2019 & \cite{Chen.2019} & User product recommendation & Dynamic preference monitoring with neural networks \\
 2019 & \cite{Kuhn.2019} & Proof of Concept & Fault location in combinatorial testing \\
 2019 & \cite{Kailkhura.2019} & Material sciences & Analogy creation and feature importance \\
 2019 & \cite{Tonekaboni.2019} & Medicine & Domain appropriate representation, \\
  & & & potential accountability, and consistency \\
 2019 & \cite{Wang.2019} & General & Psychology-based theoretical cognitive framework \\
 2019 & \cite{Cashmore.2019} & Industrial planning & Contrastive explanations \\
 2019 & \cite{Hois.2019} & Human-centric decision making & Context-based \\
 2019 & \cite{Gade.2019} & Survey for industrial application & Survey \\
 2019 & \cite{Samek.2019} & Survey & Several \\
 2019 & \cite{Gunning.2019} & Survey & Survey \\
 2019 & \cite{Holzinger.2019} & Medicine & Definition of causability and explainability \\
 2018 & \cite{Ai.2018} & Product recommendation & Knowledge-base embedding representation \\
 2018 & \cite{Bellini.2018} & (Movie) recommendations & Knowledge graphs \\ 
 2018 &  \cite{Hoffman.2018} & User acceptance & Survey \\
 2018 &  \cite{Hagras.2018} & Survey & Rule-based decisions \& labels \\
 2018 &  \cite{Adadi.2018} & Survey & Survey \\
 2018 &  \cite{Abdollahi.2018} & Health, education, justice, & survey \\
 2018 &  \cite{Holzinger.2018} & Position paper & Context-adaptive procedures \\
 2018 &  \cite{Goebel.2018} & Position paper & Bridging cognitive valley with graph information \\
 2018 & \cite{Preece.2018} & Survey & Taxonomy for \ac{xai} \\
 2017 &  \cite{Holzinger.2017} & Medicine & Hybrid distributional models \\
 \bottomrule
\end{tabular}
\vspace{0.1cm}
\end{table}
This table lists the respective work,
the domain which is discussed and the method used to explain the decision or recommendation made by the \ac{ai} method.
\textit{Reddy} discusses the requirements formulated by stakeholders for acceptance of \ac{ai} decisions in medical treatment and research while formulating the point that some argue in favour of higher accuracy algorithms instead of well-explainable ones~\cite{Reddy.2022}.
\textit{Neugebauer et al.} present a surrogate \ac{ai} model,
that addresses parameter changes of the base model and consequently highlights the relevant parameters in the decision,
aiding its explainability~\cite{Neugebauer.2021}.
\textit{Vilone and Longo} survey scientific research addressing \ac{xai} and categorise findings in human-based explanations that aim at mimicking human reasoning,
and objective metrics such as accuracy~\cite{Vilone.2021}.
\textit{Caro-Mart\'{i}nez et al.} introduce a conceptual model for e-commerce recommender systems that extends existing models with four ontological elements: User motivation and goals,
required knowledge,
the recommendation process itself,
and the presentation to the user~\cite{Caro.2021}.
\textit{Liang et al.} present a novel online \ac{ers} that,
in contrast to commonly used offline \acp{ers},
can be updated and instantly provides explanations with recommendations~\cite{Liang.2021}.
\textit{Holzinger and M\"{u}ller} discuss a mapping appraoch of explainability with causability~\cite{Holzinger.2021}.
That means creating links between the reasoning an \ac{ai} algorithm implicitely makes and the intuitive conclusions humans draw.
This is applied to the area of image-based pattern recognition in medical treatment.
\textit{Ehsan et al.} present a concept for integrating social transparency into \ac{xai} solutions~\cite{Ehsan.2021}.
They present a framework based on expert-interviews.
\textit{Angelov et al.} discuss the relation of \ac{ai} algorithms with high accuracy and high explainability factors~\cite{Angelov.2021}.
A taxonomy is provided,
global vs. local model explanation techniques for different domains and algorithms are surveyed and set in context with the remaining challenges.
\textit{Mohseni et al.} provide a survey of existing literature that clusters available methods and research approaches respective to their design goals as well as evaluation measures~\cite{Mohseni.2021}.
Their framework is founded on the distinction between the provided categories.
\textit{Belle and Papantonis} evaluate feasible methods of explainability on the use case of a data scientist that aims to convince stakeholders~\cite{Belle.2021}.
\textit{Shin} discusses the relation of causability and explainability in \ac{xai} and the influence on trust and user behaviour~\cite{Shin.2021}.
\textit{Singh et al.} evaluate methods to explain \ac{ai} conclusions in the medical domain~\cite{Singh.2020}.
\textit{Barredo Arrieta et al.} present an extensive survey on literature and solutions in \ac{xai} on which they base requirements and challenges yet to conquer~\cite{Barredo_Arrieta.2020}.
Ultimately,
they create the concept of fair \ac{ai} that explains and accounts for decisions made.
\textit{Lundberg et al.} introduce a game-theoretic model for optimal explanations in tree-based algorithms~\cite{Lundberg.2020}.
Local explanations are combined to obtain a global explanation of the trained tree in a human-understandable format.
\textit{Ploug and Holm} present the concept of contestable \ac{ai} decision making in a clinical context~\cite{Ploug.2020}.
Contestability means that the decision algorithm has to provide information to the data used, 
any system biases,
system performance in terms of algorithmic metrics,
and the decision responsibility carried by humans or algorithms.
If the decision is contested,
the algorithm has to provide insight that the alternate solution was considered as well and has been adequately taken into account.
\textit{Arya et al.} introduce a collection of explainability tools that are combined into a framework to provide researchers and data scientists with the opportunity to exctract explanations for various algorithms~\cite{Arya.2020}.
\textit{Linardatos et al.} introduce a survey and taxonomy,
distinguishing between different types of interpretability in \ac{ai} methods before presenting an exhaustive list of tools and methods~\cite{Linardatos.2020}.
\textit{Tjoa and Guan} discuss challenges and risks of explainability in medical \ac{ai} applications~\cite{Tjoa.2020}.
They survey existing solutions for different algorithm types while also introducing risks and challenges with existing solutions.
\textit{Roscher et al.} present an overview of methods to conserve scientific interpretability and explainability in natural sciences~\cite{Roscher.2020}.
\textit{Beaudoin et al.} introduce a framework for explainability that can be applied in multidisciplinary settings~\cite{Beaudouin.2020}.
Three steps are required to obtain the suitable method of explainability:
First,
define the contextual factors regarding the explanation.
Second,
analyse the tools available for the technical problem at hand and third,
chose the suitable global and local explanation tools that are compared to seven factors of cost created by lacking explanation.
\textit{Coeckelbergh} discusses the ethical and philosophical responsibility of decisions made by \ac{ai} algorithms and evaluates the responsibility of agents for their decisions in time and conditional dependencies~\cite{Coeckelbergh.2020}.
\textit{Bhatt et al.} introduce a framework based on findings of a study that investigates the target audience of explainability in \ac{ai}~\cite{Bhatt.2020}.
They find that most explainability approaches are created for machine learning engineers to adapt their model.
Consequently,
they propose a framework that allows the choice of a target audience and adapts the explainability accordingly.
\textit{Ammar and Shaban-Nejad} present a recommendation system for mental health analysis that is based on ontological knowledge of the domain~\cite{Ammar.2020}.
\textit{Zhang and Chen} survey \ac{ers} and their uses for end-users as well as developers~\cite{Zhang.2020}.
\textit{Confalonieri et al.} discuss the historical implications of \ac{xai} and propose criteria for explanations deemed necessary for human understanding of explanations~\cite{Confalonieri.2020}.
\textit{Amann et al.} evaluate the legal, 
ethical and organisational aspects of \ac{ai} decisions in the medical domain~\cite{Amann.2020}.
\textit{Chen et al.} develop a dynamic \ac{ers} that monitors user preferences and maps them to aspects of a product review to increase recommendation quality~\cite{Chen.2019}.
\textit{Kuhn and Kacker} apply the well-established problem of fault location in combinatorial testing,
where a software fails only after input of several bogus values that have to be identified,
to the explainability problem in \ac{ai}~\cite{Kuhn.2019}.
Based on feature combinations,
the importance of any given feature for the decision is derived and thus used to explain the decision making process.
\textit{Kailkhura et al.} present a methodology for explaining classifications of \ac{ai} algorithms in the domain of chemical material sciences~\cite{Kailkhura.2019}.
Well-known and established algorithms,
such as XGBoost~\cite{Chen.2016},
are extended with explanations regarding analogies in decisions to mimic human reasoning as well as feature importance to provide insight into the model design.
According to their methodology,
an \ac{ai} model used for scientific research has to be transparent,
the output should be interpretable,
and the scientific result has to be explainable.
Several methods are surveyed with respect to their dimensions of interpretability and the required integration of domain knowledge into the models.
\textit{Tonekaboni et al.} discuss requirements and conditions for \ac{ai} in medical studies~\cite{Tonekaboni.2019}.
Explainability in this context describes the justifiability of results to stakeholders and colleagues, 
meaning explanations need to take stakeholder interests into account.
The three metrics identified in their work contain domain appropriate representation,
meaning that different disciplines in the medical field require different sets of information.
Second,
potential accountability describes a workflow for follow-up checking of the model outcome,
meaning in the further treatment,
the model decision should be validated with additional lab tests or checking in on patients.
Third,
consistency means that results provided by \ac{ai} models should only be based on clinical variables and that any change in variable,
and consequently outcome,
should encompass a different but fitting explanation.
\textit{Wang et al.} apply psychology and cognitive sciences to obtain an understanding of human bias and interpretation~\cite{Wang.2019}. Based on this information,
they extract information from different decision algorithms and collect them in a framework where the individual data points are presented combined with their relevance for a decision.
\textit{Cashmore et al.} discuss explainability as a service~\cite{Cashmore.2019}.
In an industrial planning scenario,
the user is asked to provide alternative plans with constraints from which an automated planner derives constraints that are the basis for an automated plan.
These constraints are used to justify the automatatically generated plan.
\textit{Hois et al.} analyse the need for human-centric \ac{ai}-based decision making and the subsequent context-dependant explanation~\cite{Hois.2019}.
Humans as users are the main focus of the explanation that has to be tailored to situation and context.
\textit{Gade et al.} introduce challenges and existing solutions for the application of explainability in various domains~\cite{Gade.2019}.
\textit{Samek et al.} dedicate a book to an overview of challenges,
solutions,
and visualisation of \ac{xai}~\cite{Samek.2019}.
\textit{Gunning et al.} summarise challenges and user expectations on \ac{xai}~\cite{Gunning.2019}.
\textit{Holzinger et al.} evaluate causability as the property of a person in terms of understanding automated reasoning and compare it to explainability as the property of a model to provide reasoning in histopatological examples~\cite{Holzinger.2019}.
\textit{Ai et al.} present a deep learning approach on knowledge bases for recommender systems to improve the explainability aspects of recommendations~\cite{Ai.2018}.
\textit{Bellini et al.} introduce a knowledge-graph-based method to explain user recommendations,
thus increasing user satisfaction and acceptance~\cite{Bellini.2018}.
\textit{Hoffman et al.} present metrics regarding the acceptance and satisfaction of users regarding the explainability in different \ac{xai} methods~\cite{Hoffman.2018}.
\textit{Hagras} discusses the challenges in \ac{xai} with a focus on dimensions of explainability as well as mapping of output to interpretation~\cite{Hagras.2018}.
They discuss the necessity of rule-based mapping of decisions to explanations as well as labels and introduce fuzzy logic systems that can be used to explain class affiliation in a human-understandable fashion.
\textit{Adadi and Berrada} introduce a survey of concepts and challenges in \ac{xai}~\cite{Adadi.2018}.
\textit{Abdollahi and Nasraoui} evaluate the fairness of \acp{ers} in the context of health,
justice,
education,
and criminal investigations~\cite{Abdollahi.2018}. 
\textit{Holzinger} provides a positional paper discussing the need for context-adaptive procedures in \ac{ai} in order to make decisions understandable for humans~\cite{Holzinger.2018}.
\textit{Goebel et al.} discuss the need for \ac{xai} in the context of increasingly complex \ac{ai} systems~\cite{Goebel.2018}.
Interactions between data points that can be represented in a graph can function as vectors of explainability,
as these areas with high importance can be identified.
\textit{Preece} provides a survey and taxonomy of challenges and a possible solution of \ac{xai}~\cite{Preece.2018}.
\textit{Holzinger et al.} present research on explainable \ac{ai} in the medical domain~\cite{Holzinger.2017}.
They argue that benefits can be obtained from \ac{ai} algorithms,
but only in case the results are explainable and can be argumented.
They summarise that hybrid distributional models which map sparse graph-based information to dense vectors of information are suitable to provide context by linking this information to lexical sources and knowledge bases.

\section{What is Explainability of AI}
\label{sec:explainability_in_ai}
\begin{figure}[ht]
	\centering
    \includegraphics[width=0.9\textwidth]{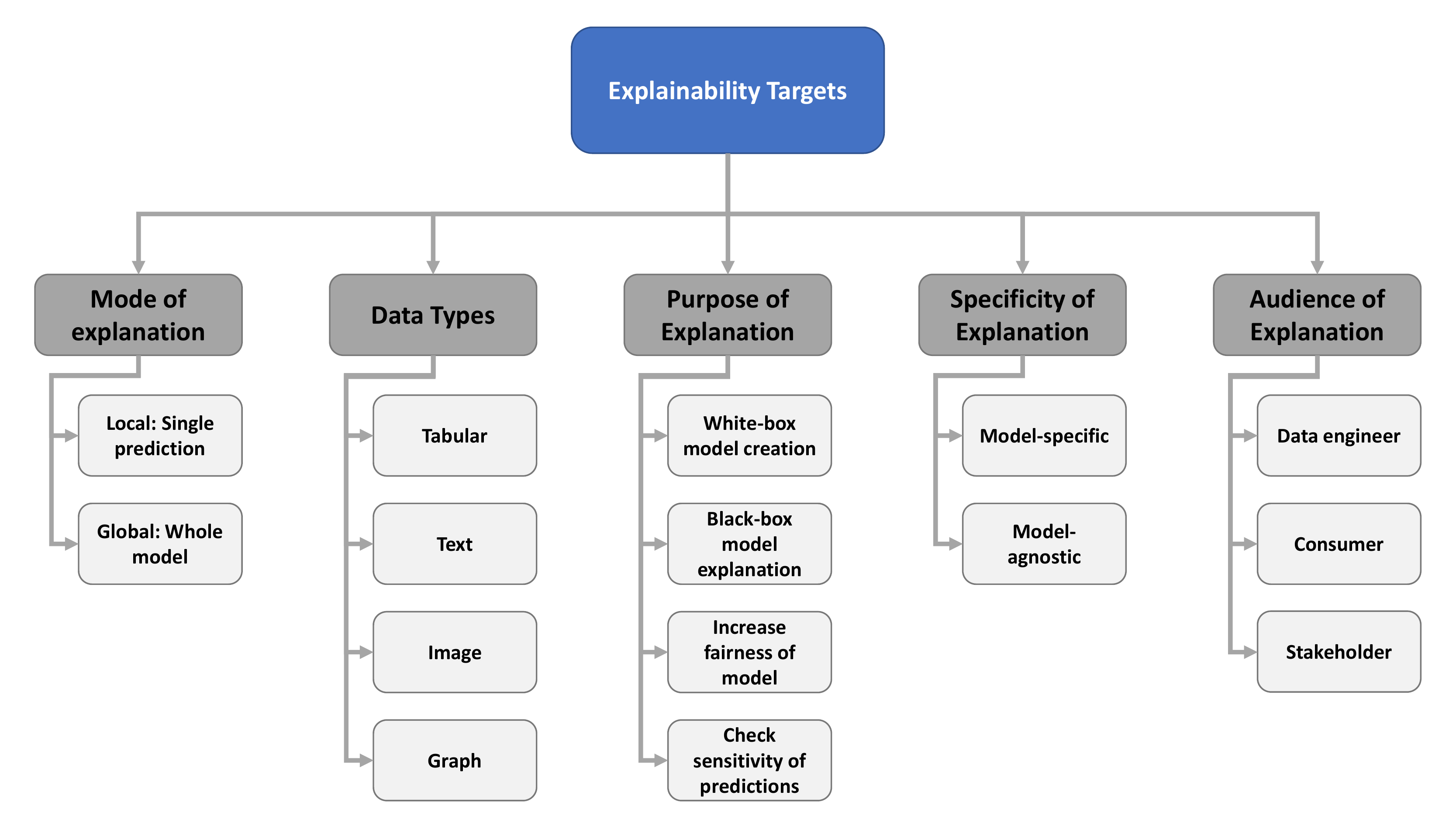}
	\caption{Explainability Goals According to \textit{Linardatos et. al}~\cite{Linardatos.2020}, Extended}
	\label{fig:expl_goals}
\end{figure}

According to \textit{Linardatos et al.}, 
explainability or interpretability of \ac{ai} algorithms depends on four dimensions~\cite{Linardatos.2020}:
\begin{itemize}
    \item Local vs global,
    \item data types that are processed by the algorithm,
    \item the purpose of interpretability, i.e. when and how should the model be explained, and
    \item the generalisation of the interpretation, i.e. can the explanation be applied to individual models or to any model.
\end{itemize}
Furthermore,
several works address the issue of target audience~\cite{Zhang.2020},
which should also be taken into consideration.
These dimensions are shown in Figure~\ref{fig:expl_goals}.
A trend that is observed by every work surveyed in this paper is the relationship between complexity of algorithms and difficulty of explainability.
Models based on Decision Trees,
such as Random Forests~\cite{Cutler.2012} and XGBoost~\cite{Chen.2016} not only perform well in praxis~\cite{Duque_Anton.2019},
but also allow for a good understanding of global explainability.
That means the model provides information about the relevance of its learned features.
Furthermore,
the splits allow insight into the creation of output.
In contrast,
\acp{dnn} create complex models that cannot easily be understood by humans.
In case of image and video processing, 
methods are available to highlight the areas in an image that induce the highest neuron output and thus are most relevant for the algorithmic output~\cite{Singh.2020}.
This method cannot be applied as easily to abstract,
high-dimensional information which is inherently difficult to interpret for humans.
Several works use contextual information as well as knowledge bases that provide additional information and allow the generation and connection of the original algorithm with algorithms used to explain the outcome~\cite{Caro.2021,Ehsan.2021}.
This can be closely combined with understanding of how humans explain decisions and understand explanations~\cite{Wang.2019}.\par
This question is closely coupled to the target audience of explanations:
while data scientists and engineers are interested in model performance,
users might be more interested in the base information that led to a certain outcome.
In domain-specific applications,
the explainability might be used to increase customer satisfaction and thus revenue~\cite{Caro.2021},
meaning the business operators need to understand not only what customers desire,
but why.
In medical applications,
compliance and liability play important roles~\cite{Holzinger.2021,Tjoa.2020,Singh.2020},
meaning someone has to be responsible for a decision.
Consequently,
a medical professional has to be able to obtain information of the root cause of an automated decision.
In social and judical applications,
fairness independant of human-induced bias is relevant as well as argumentation regarding the reasons for conclusions~\cite{Ehsan.2021,Abdollahi.2018}. \par
This section shows that explainability is a broad and rather abstract concept,
once one is trying to implement it.
Not only does the data and algorithm have a strong influence on the methods of explainability that are possible in the first place,
also the goal and audience of an explanation are relevant.
The domains and consequentially the goals of explainability are discussed in the next section.

\section{Domains for Expainability}
\label{sec:ai_scenarios}
It is noteworthy that there is distinction in related work,
as discussed in Section~\ref{sec:related_work},
regarding the use cases discussed:
The first part of papers present solutions,
and sometimes challenges, to enhance explainability of certain algorithms or types of algorithms.
At the same time,
the second part of papers discusses requirements,
challenges and solutions of certain fields or domains,
without specific algorithms in mind.
Furthermore,
several papers address \ac{xai} as a means for improved outcomes and stakeholder or user acceptance,
while others solely address technical challenges.
For a holistic explainability architecture,
those have to be merged.
Generally,
the two different ways of approaching explainability show in this survey.
Domain experts have a concrete challenge that can be solved,
or improved,
with \ac{ai} methods.
Alternatively,
an \ac{ai} method is already solving the challenge and the domain expert needs reasoning for the model,
be it to appease stakeholders or to increase customer satisfaction.
Such approaches often use knowledge bases of methods to solve the issue without \ac{ai}. \\ \par
The other approach is from machine learning and \ac{ai} experts who aim to make their models more transparent.
Often,
this is based in the need to understand the model in order to increase performance,
resulting in explainability by accuracy and sensitivity metrics.
Such information is difficult to interpret by non-experts and thus not suited for explainability for users.
In summary,
the different domains for explainability highlight the need for distinction in the solution.
In recommender systems for e-commerce,
a wrong recommendation will not have severe consequences,
in stark contrast to recommendation systems for medical examination.
Understanding models to fine-tune them is a valid task for data scientists and machine learning experts,
while a ontological explanation similar to human reasoning is necessary for integration of \ac{ai} models into socio-economic and organisational solutions.
While it is a highly interesting concept,
a holistic solution to explainability does not seem likely.

\section{Conclusion}
\label{sec:conclusion}
A large body of research regarding \ac{xai} has been presented in this work.
This body of research spans several domains,
including medicine and healthcare,
commerce,
criminal and social sciences,
and recommender systems.
While highlighting that all of these domains face similar challenges,
the mass of surveys and taxonomies discussed in this work also show the wide span of requirements. 
Additionally,
the technique of explainability strongly depends on the employed \ac{ai} algorithm.
Simpler algorithms,
for example tree-based ones,
can be understood by humans relatively easy as the information of the decision can be extracted directly from the model in a semantically understandable fashion.
More complex algorithms,
such as \acp{dnn} prove to be more difficult.
If the data they analyse can be understood intuitively by humans,
as is the case in image processing,
the neuronal activity can be displayed.
However,
in complex and unstructured data,
such as high dimensional tables,
highlighting of the values does not provide humans with an understanding of the reasoning.
Here,
knowledge bases to map against or ontological translations of the results are required in order to provide an understanding. \\ \par
\ac{ai}-based recommender systems are bound to drastically increase performance and quality of services they provide.
Due to the variety of said services,
individual explanation frameworks will be applied.
Furthermore,
recipient and reason for wanting an explanation play an important role in chosing an explanation framework.
In the future,
\ac{ai} systems might increasingly make decisions autonomously,
i.e. without supervision and approval of a human operator.
Such applications will require strict regulation and liability frameworks to ensure that \ac{ai} methods perform in an expected and sound manner.

\section*{Acknowledgement}
This work has been supported by the Federal Ministry of Education and Research (BMBF) of the Federal Republic of Germany (Foerderkennzeichen 19I21028R, WaVe) and the Investment and Structure Bank (ISB) Rhineland Palatinate (Foerderkennzeichen P1SZ26, InnoTop).
The authors alone are responsible for the content of this paper.

\bibliographystyle{splncs03}
\bibliography{bibliography}

\end{document}